\documentclass[lettersize,journal]{IEEEtran}
\usepackage{amsmath,amsfonts}
\usepackage{algorithmic}
\usepackage{array}
\usepackage[caption=false,font=normalsize,labelfont=sf,textfont=sf]{subfig}
\usepackage{textcomp}
\usepackage{stfloats}
\usepackage{url}
\usepackage{cite}
\usepackage{verbatim}
\usepackage{graphicx}
\usepackage{multirow}
\usepackage{xcolor,colortbl}
\def\BibTeX{{\rm B\kern-.05em{\sc i\kern-.025em b}\kern-.08em
    T\kern-.1667em\lower.7ex\hbox{E}\kern-.125emX}}
\usepackage{balance}
\begin{document}
\title{CINFormer: Transformer network with \par multi-stage CNN feature injection for \par surface defect segmentation}
\author{Xiaoheng Jiang, Kaiyi Guo, Yang Lu, Feng Yan, Hao Liu, Jiale Cao \par 
Mingliang Xu, and Dacheng Tao, Fellow, IEEE
\thanks{Manuscript created August 2023; This work was supported in part by the Nation Key Research and Development Program of China under Grant 2021YFB3301500; in part by the National Natural Science Foundation of China under Grant 62172371, U21B2037, 62102370, 62272421; in part by Natural Science Foundation of Henan Province under Grant 232300421093 and the Foundation for University Key Research of Henan Province under Grant 21A520040; in part by CAAI-Huawei MindSpore OpenFund. (\emph{Corresponding author: Mingliang Xu})

Xiaoheng Jiang, Yang Lu, Hao Liu, and Mingliang Xu are with School of Computer Science and Artificial Intelligence, Zhengzhou University, Zhengzhou, China; Engineering Research Center of Intelligent Swarm Systems, Ministry of Education, Zhengzhou, China; National Supercomputing Center in Zhengzhou, Zhengzhou, China (e-mail: jiangxiaoheng@zzu.edu.cn, ieylu@zzu.edu.cn, HaoLiu1989@hotmail.com, iexumingliang@zzu.edu.cn)

Kaiyi Guo and Feng Yan are with School of Computer Science and Artificial Intelligence, Zhengzhou University, Zhengzhou, China (e-mail: gky10523@gmail.com, ieyanfeng@163.com)

Jiale Cao is with School of Electrical and Information Engineering, Tianjin University, Tianjin, China (e-mail: connor@tju.edu.cn)

Dacheng Tao is with the Sydney AI Centre and the School of Computer Science, Faculty of Engineering, The University of Sydney, Darlington, NSW 2008, Australia (e-mail: dacheng.tao@gmail.com)

}}

\markboth{IEEE Transactions on Neural Networks and Learning Systems, ~Vol.~xx, No.~xx, xx~2023}%
{How to Use the IEEEtran \LaTeX \ Templates}

\maketitle

\begin{abstract}
Surface defect inspection is of great importance for industrial manufacture and production. Though defect inspection methods based on deep learning have made significant progress, there are still some challenges for these methods, such as indistinguishable weak defects and defect-like interference in the background. 
To address these issues, we propose a transformer network with multi-stage CNN (Convolutional Neural Network) feature injection for surface defect segmentation, which is a UNet-like structure named CINFormer. 
CINFormer presents a simple yet effective feature integration mechanism that injects the multi-level CNN features of the input image into different stages of the transformer network in the encoder. This can maintain the merit of CNN capturing detailed features and that of transformer depressing noises in the background, which facilitates accurate defect detection.
In addition, CINFormer presents a Top-K self-attention module to focus on tokens with more important information about the defects, so as to further reduce the impact of the redundant background. 
Extensive experiments conducted on the surface defect datasets  DAGM 2007, Magnetic tile, and NEU show that the proposed CINFormer achieves state-of-the-art performance in defect detection. 
\end{abstract}


\section{Introduction}
\IEEEPARstart{D}{efect} inspection~\cite{bergmann2019mvtec, liu2021anomaly} is an important task in industrial automatic production. It aims to detect anomalous or defective areas on the surface of the products. Traditional manual defect inspection methods rely on experienced workers and suffer from low efficiency and poor accuracy.

With the development of deep learning~\cite{rumelhart1986learning,krizhevsky2012imagenet}, 
deep convolution neural networks (CNNs) have achieved great success in many computer vision tasks such as image classification~\cite{he2016deep,tan2019efficientnet,simonyan2014very}, object detection~\cite{tang2021qbox,li2021looking,9585030,9525042}, and semantic segmentation~\cite{wang2018understanding,zhang2021document}.
%
Therefore, many CNN-based defect inspection methods~\cite{tabernik2020segmentation,he2019end} have sprung up. Among them, pixel-level defect detection methods can provide fine-grained information about defects. Most of these methods
~\cite{fang2022tactile,dong2019pga,wei2020detecting,su2021rcag} introduce attention mechanisms into models to improve the defect detection ability. However, these methods can not accurately detect the defects in complex scenes with noise interference, especially when defects present a weak appearance. 
%

\begin{figure}[t]
    \centering
    \includegraphics[width=\linewidth]{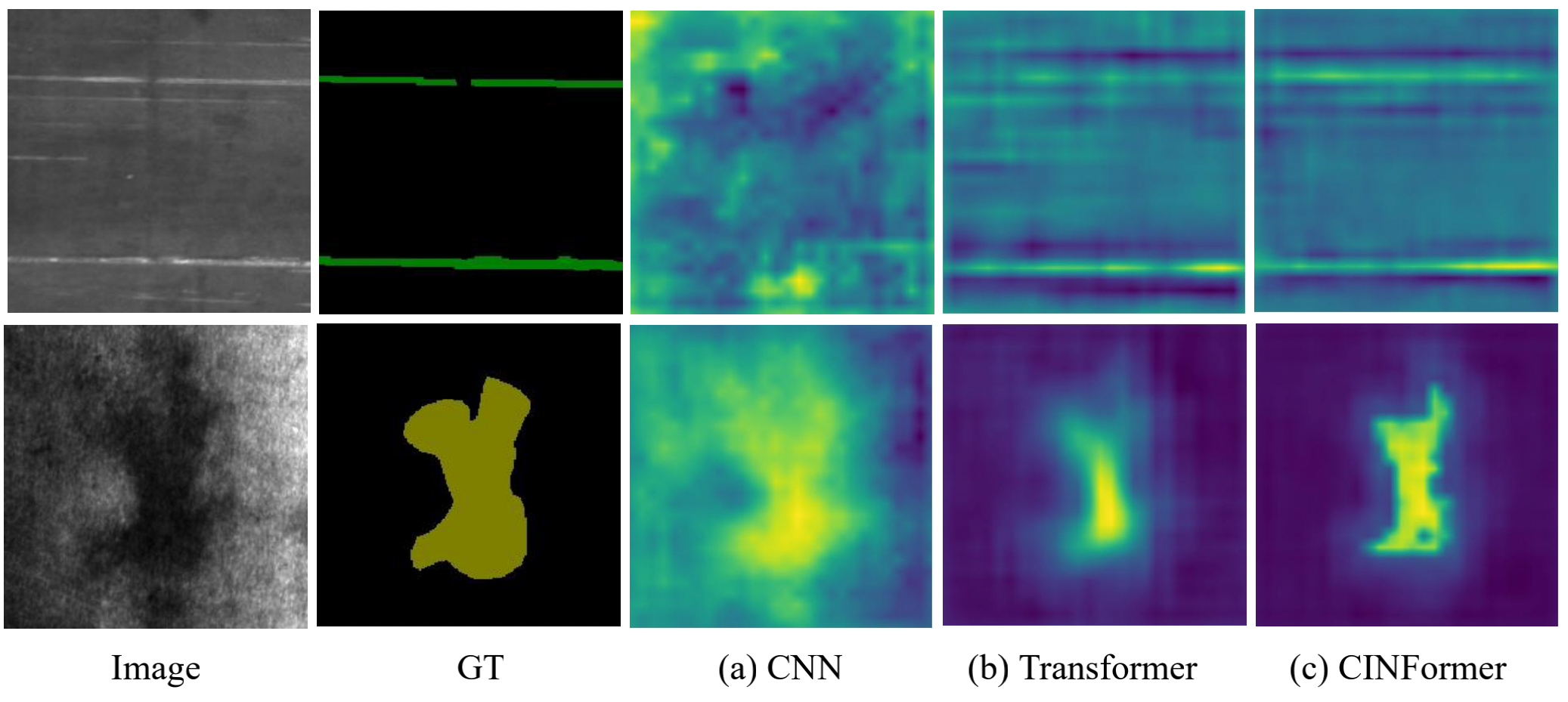}
    \caption{ Comparison of feature visualization for CNN (a), transformer (b), and the proposed CINFormer (c). It is noted that the features are from the last stage of the corresponding models. It can be observed that CINFormer can better focus on defect areas and suppress redundant background interference.
    }
    \label{fig1}
\end{figure} 

Recently, transformers~\cite{vaswani2017attention} have achieved excellent performance in the natural language processing task. And with the advent of ViT ~\cite{dosovitskiy2020image}, the transformer shows great potential in computer vision, which can model long-distance dependencies through self-attention module. 
However, it is difficult for existing transformer models to accurately capture detailed defect information since the self-attention module is essentially a low-pass filter~\cite{park2022vision}. As a result, it tends to filter out high-frequency signals, which results in losing the detailed information of defects, especially for those weak defects, as presented in Fig.~\ref{fig1}.

To address these problems, some works~\cite{peng2021conformer,li2022uniformer,wu2021cvt} combine CNN features and transformer features to strengthen the representation ability of features. 
Some other methods~\cite{liu2021swin,qin2022cosformer,wang2021pyramid} introduce local properties of convolution into the transformer model ~\cite{liu2021swin,qin2022cosformer,wang2021pyramid} to better learn local features.
In this way, the model can capture detailed information while suppressing background interference.
Although these methods make the transformer better capture detailed information, they still confront challenges caused by weak defects and complex backgrounds.

This paper aims to design a cooperation strategy that can effectively combine the CNN network and the transformer network for surface defect segmentation. To this end, we propose a UNet-like transformer network named CINFormer.  
Specifically, a CNN stem is used to generate multi-level convolutional features which are injected into different stages of the transformer in the encoder of CINFormer. 
It is noted that the feature injection is a one-way procedure from CNN to the transformer. This can make CINFormer better absorb rich original detailed defect features while maintaining the ability of suppressing background noise. 
Furthermore, a Top-K self-attention module is further proposed to remedy the situation that detailed defect features can be drowned out to some extent by the background information.
It selects more meaningful tokens and channels by ranking their variances to calculate attention. This can alleviate the impact of redundant background information, which facilitates the detection of weak defects.

It should be noted that the proposed injection strategy is inspired by Conformer~\cite{peng2021conformer} but differs from it. Conformer is a bidirectional feature injection pattern with features of CNN and transformer interacting with each other. Though this strategy can promote the full merging of features from the transformer and CNN, it at the same time compromises the ability of the CNN branch to represent the detailed features of the weak defects. In contrast, our proposed one-way multi-stage CNN feature injection method can effectively retain detailed CNN features and make them fully merged with transformer features, which facilitates defect detection. 
In addition, Conformer needs to train the CNN and Transformer branches simultaneously. Different from it, our CINFormer can just fine-tune the Transformer part while keeping the CNN part fixed. Even so, CINFormer outperforms Conformer on the defect datasets. When trained in an end-to-end way, CINFormer obtains a better performance.

In summary, our contributions are as follows:
\begin{enumerate}
\item  We propose a UNet-like defect segmentation network called CINFormer that combines CNN and transformer as the encoder. The multi-level CNN features are injected into different stages at the transformer network to strengthen the representation capability of the features.

\item  We present a Top-K self-attention module to alleviate the impact of the redundant background information.  It helps focus on essential defect features while suppressing background interference.

\item  The extensive experiments on three typical surface defect datasets demonstrate that the proposed method is effective in different defect scenes.
\end{enumerate}

\section{Related Work}

\begin{figure*}[t]
    \centering
    \includegraphics[width=\linewidth]{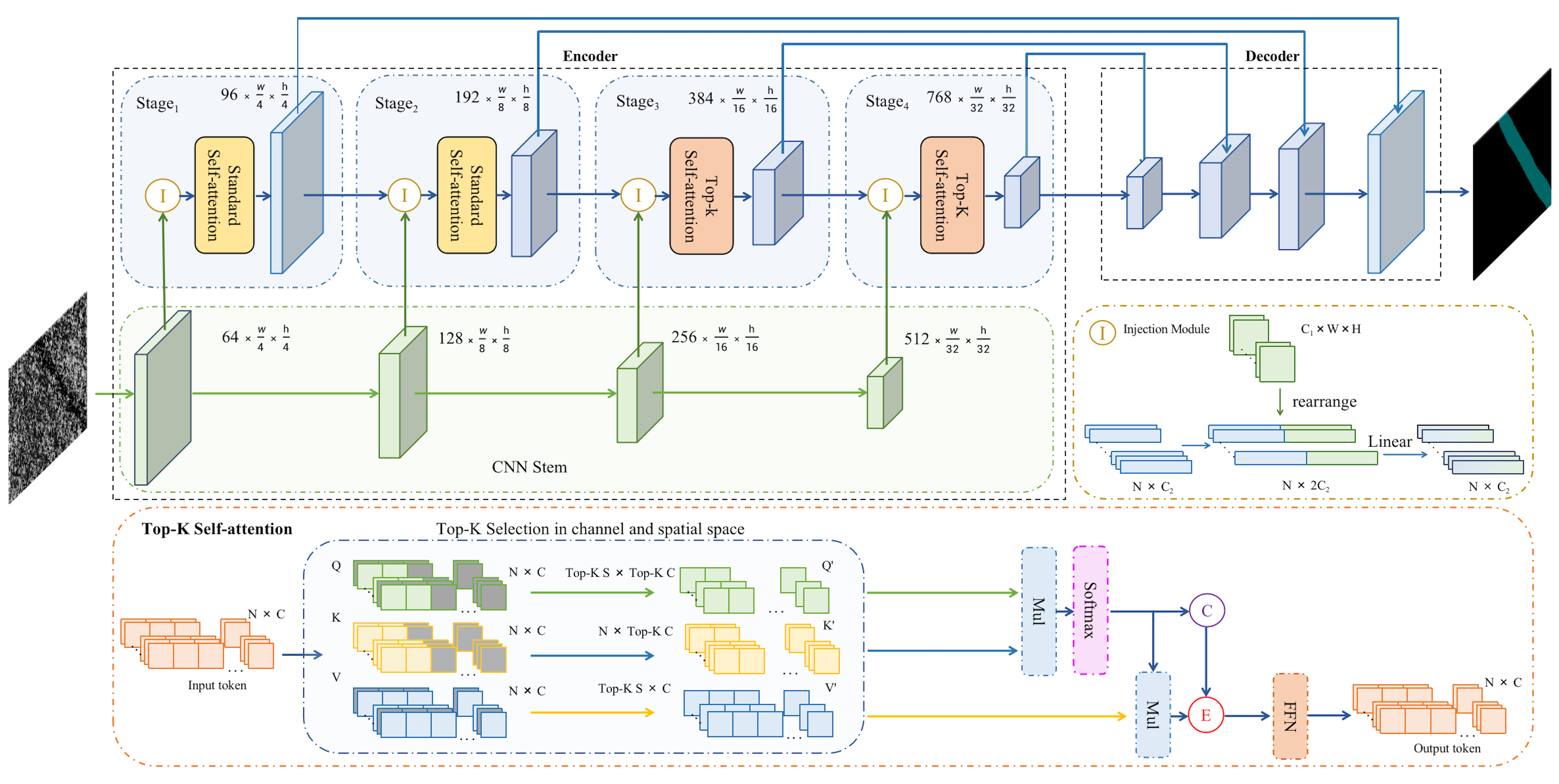}
    \caption{The architecture of the CINFormer. The CINFormer is a UNet-like encoder-decoder network. The encoder is a transformer network with multi-level convolutional features injected into its different stages. In addition, the Top-K self-attention module is introduced into the transformer block to reduce the impact of redundant background information. It selects the top $k$ most important tokens and channels to retain the most useful information while suppressing the redundant background information.
    }
    \label{fig2}
\end{figure*}

\subsection{CNN-based Defect Inspection}
Traditional machine visions utilize low-level texture features to detect defects. These methods can not effectively defect complex defects with low contrast appearance and small size.
Recently, with the rapid development of deep learning, some CNN-based methods have been widely applied to surface defect detection. 
For instance, Tabernik et al.~\cite{tabernik2020segmentation} proposed a two-stage network, which introduces a segmentation subnetwork in the classification network. 
%
%
Cui et al.~\cite{yang2020deep} integrate fine-grained details of preceding layers at each deep layer, which facilitates the detection of small objects. 

However, it is hard for these methods to identify defects in complex scenes.  Benefiting from the attention mechanism that highlights important features, some methods introduce it into the model to strengthen the feature representation.
Based on the human visual gain mechanism, Wei et al.~\cite{wei2020detecting} constructed a Faster VG-RCNN that integrates attention-related visual gain mechanism to better detect small defects.
Su et al.~\cite{su2021baf} integrate the improved multi-head nonlocal self-attention into FPN to highlight defect features. 
Yang et al.\cite{yang2023pixel} introduce the spatial attention module into each residual unit of the backbone to make the network adaptively focus on defect regions.
%
Similarly, Jiang et al.~\cite{9933904} use joint channel-spatial attention weights to selectively fuse high-level and low-level features.
Xiang ea al. \cite{xiang2023multi} introduce the attention module after high-level and low-level feature fusion to strengthen feature representation.
%
%
However, these CNN-based methods are still challenged by defects in complex scenes because CNN features are vulnerable to background interference.


\subsection{Vision Transformer}
Recently, Dosovitskiy et al.~\cite{dosovitskiy2020image} proposed a vision transformer, which shows the strong ability to capture global features benefiting from the self-attention mechanism. 
To deal with variations in the scale of objects, Liu et al.~\cite{liu2021swin} proposed a hierarchical transformer based on window self-attention.
Wang et al.~\cite{wang2021crossformer} proposed a CrossFormer to improve Transformer's ability to build cross-scale interactions.
Xie et al.~\cite{xie2021segformer} constructed a hierarchical transformer as an encoder to generate multi-level features, and fused these features through MLP layers in the decoder. 
These methods focus on drawing global relationships, which are robust against background interference.
However, these methods are prone to lose some detailed information, Because it does not pay attention to local information. 

To deal with the above problem, Liu et al.~\cite{liu2021swin} proposed a local window mechanism to perform self-attention operations within the window. 
Yang et al.~\cite{yang2021focal} proposed a Focal Self-Attention (FSA), which pays fine-grained attention to the area around the current token and coarse-grained attention to the area far away from the current token. In this way, it can more effectively capture local and global attention. 
Xia et al.~\cite{xia2022vision} proposed a deformable self-attention module in which key and value in self-attention are sampled to flexibly learn informative features. 
Fan et al.~\cite{Fan_2021_ICCV} proposed a multiscale Transformer, which creates a multiscale feature pyramid that learns low-level information at early layers and semantic information at deep layers.
Ren et al.~\cite{ren2022shunted} proposed a Shunted Self-Attention (SSA), which can capture multi-scale features through multi-scale token aggregation. 
However, these methods still do not solve the problem that self-attention is low-pass filtering, making it difficult for them to deal with weak or minor defects.

\subsection{Combination of Transformer and CNN}
Due to the local properties of CNN, it is prone to bring in noises when dealing with defects in complex scenes. In contrast, the transformer model can draw long-distance dependencies and suppress noise, while it is weak in capturing detailed information about objects. 
Therefore, some methods improve the networks' performance by combining CNN and transformer networks. For example,
Peng et al.~\cite{peng2021conformer} designed a hybrid network structure termed Conformer, which enhances representation learning by combining local and global features in an interactive way.
Wu et al.~\cite{wu2021cvt} use convolutional projection instead of linear projection in each self-attention block, which can make the transformer model capture local details.
Xie et al.~\cite{xie2022pyramid} proposed a novel one-stage framework called Pyramid Grafting Network (PGNet), which utilizes global information to strengthen CNN's feature representation in the decoder.
Li et al.~\cite{li2022uniformer} designed a novel Unified transformer (UniFormer) to learn both local and global token affinity by integrating 3D convolution and spatiotemporal self-attention.
Although these methods inherit the local properties of CNNs while maintaining the global merits of transformers, they still confront challenges posed by weak defects and complex background.

%
%
Different from the above methods, the proposed CINformer presents a one-way multi-stage feature injection method from the CNN to the transformer network, which can better preserve the local details of the CNN. Furthermore, a Top-K self-attention module is proposed to make the model focus on defect information by selecting important tokens or channels to calculate self-attention.

\section{Proposed Method}

We briefly introduce the overall architecture of the model in Section III-A. Then the transformer with CNN injection is described in Section III-B. Finally, the proposed Top-K self-attention module is described in Section III-C.

\subsection{Overall Architecture}
As depicted in Fig.~\ref{fig2}, the proposed network is constructed based on a U-shaped architecture.
A multi-level transformer network is adopted as the encoder in which each level is injected by CNN features.
Specifically, we adopt the ResNet-18~\cite{he2016deep} and the Swin-T~\cite{liu2021swin} as the CNN and transformer backbones, respectively.
Based on the ResNet-18, a CNN stem is constructed to generate hierarchical CNN features, denoted as \{${R_i|i = 1, 2, 3, 4}$\}.
It is noted that the transformer network takes the convolutional feature $R_1$ instead of images as the input.
The other convolutional features $R_2$, $R_3$, and $R_4$ are injected into features of different stages of the transformer network, which can provide the transformer with detailed defect features. 
The detailed CNN features are beneficial for segmenting those small defects.

The decoder consists of four decoding stages, each of which fuses high-level features and the corresponding low-level features in the encoder.
Besides, a Top-K self-attention module is presented to alleviate the impact of redundant background information. The proposed Top-K self-attention module selects important tokens or channels to calculate self-attention, which is used in the last two stages of the transformer network to help the model attend to important defect regions.

\subsection{Transformer with CNN Injection}
The illustration of the transformer with CNN injection is shown in Fig.~\ref{fig2}. In CNN, features at deep layers tend to contain rich semantic information about defects, while features at shallow layers contain abundant detailed information about defects.
So we construct a CNN stem through FPN \cite{lin2017feature} to generate multi-level features. The FPN fuses high-level and low-level features from top to down. In FPN, high-level features are upsampled and concatenated with the low-level features, and fed into 3$\times$3 convolutions for further fusion.
Compared with the multi-level features of the CNN backbone, the multi-level features generated by FPN contain richer semantic information and detailed information. So we use the multi-level features \{${R_i|i = 1, 2, 3, 4}$\} in the top-down path as the features injected into Transformer.

The convolutional feature $R_1$ is taken as the input of the transformer network, generating transformer features \{${S_i|i = 1, 2, 3, 4}$\} of different stages. The other features are injected into the subsequent transformer stages, respectively. Each transformer stage takes the fused features of transformer features and convolution features as input.
Specifically, the convolutional feature $R_i$ is applied a $1\times 1$ convolution to adjust the number of channels. The transformed convolutional feature is reshaped from $\mathbb{R}^{C\times H \times W}$ into $\mathbb{R}^{C\times N}$, where $N=H \times W$. The reshaped feature is fused with transformer feature $S_{i-1}$ and fed into transformer stage $i$. Mathematically, the injection process can be described as follows:
\begin{gather}
    R'_i = Reshape(Conv(R_i)) \\
    Y = Linear(Concat(S_{i-1}, R'_i))
\end{gather}
where $Reshape(.)$ represents reshape operation. $Concat(.)$ represents concatenation operation. $Conv(.)$ denotes the convolution operation. And the $Linear(.)$ denotes linear projection.

\subsection{Top-K Self-attention Module}
\begin{figure}[t]
    \centering
    \includegraphics[width=3in]{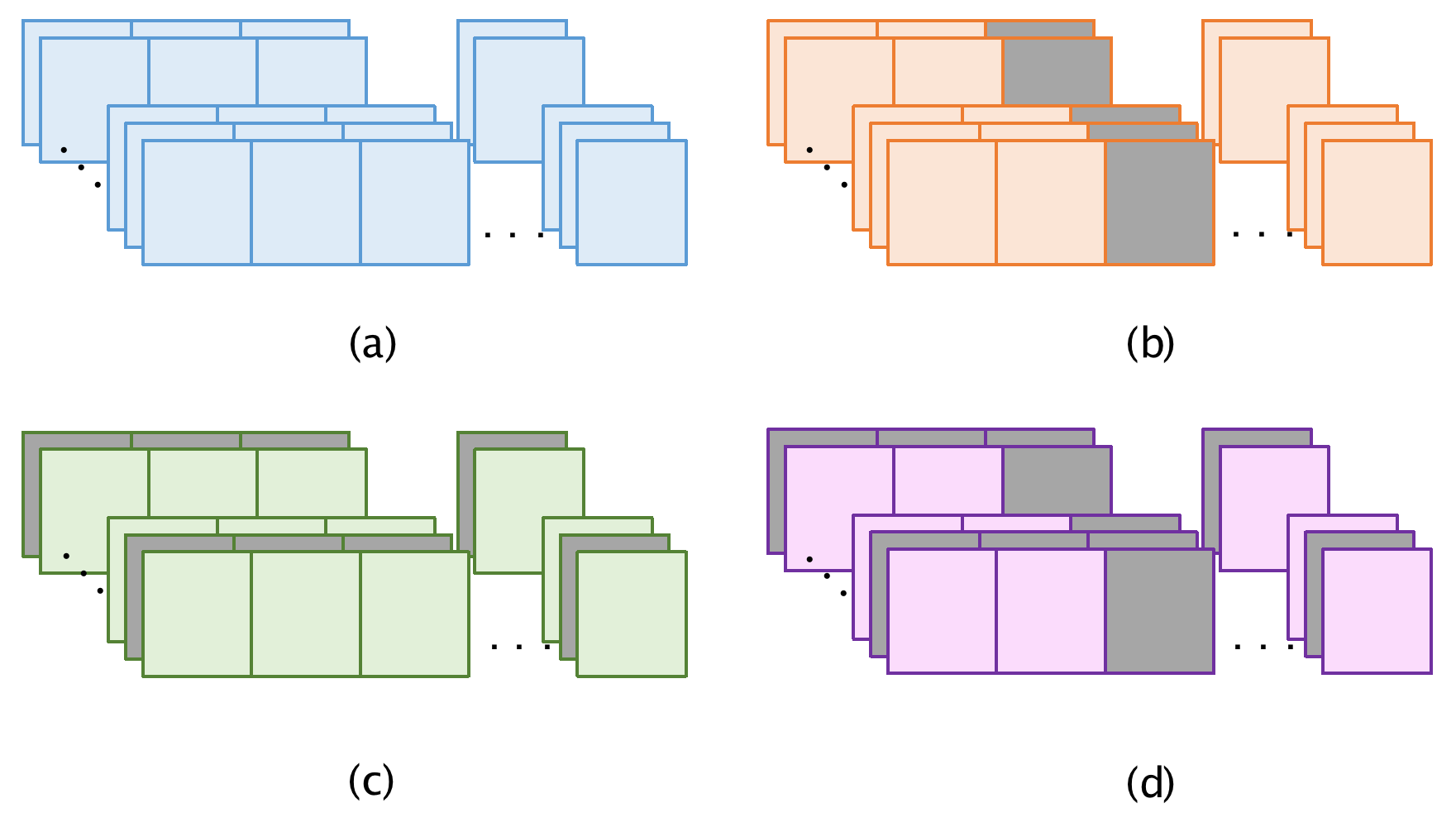}
    \caption{The illustration of the Top-K self-attention mechanism. (a) $\sim$ (d) represents tokens without selection, tokens with token selection, tokens with channel selection, and tokens with both token and channel selection, respectively. The unselected tokens and channels are represented in gray.}
    \label{fig3}
\end{figure}

Traditional self-attention computes the feature similarity between any two spatial locations. As a result, this brings redundant information, which is likely to drown out some detailed defect information. 
This is because the defects are usually weak and small, which are easily confused by the background. Traditional self-attention utilizes all tokens to produce the attention map, which brings in redundant background information. 
To remedy this problem, we present a simple but effective Top-K self-attention module, which is used in the last two transformer stages. The proposed self-attention module selects the top $k$ important tokens and channels to compute the attention. As a result, it can highlight important defect features and filter out some redundant background information. At the same time, it brings lower computational complexity than the self-attention module. The details of the proposed self-attention module are shown in Fig.~\ref{fig3}. 

We will present the proof by visualizing the feature maps of the original self-attention and Top-K self-attention module in Fig.~\ref{fig7} in our paper. The feature map of the Top-K self-attention can focus on defect information while suppressing redundant background information compared with that of the original self-attention.
In our module, given input feature $X \in \mathbb{R}^{N \times C}$, it is linearly mapped into queries \textbf{Q}, keys \textbf{K}, and values \textbf{V}. Mathematically, we have:
\begin{gather}
    \textbf{Q}=XW_q, \textbf{K}=XW_k, \textbf{V}= XW_v
\end{gather}
where $W_q$, $W_k$ and $W_v$ represent three matrices of $\mathbb{R}^{C\times C}$.

With \textbf{K}, \textbf{Q}, and \textbf{V} obtained, we use the Top-K self-attention mechanism to select the top $k$ most important tokens and channels to keep the most useful information and suppress the redundant background information. 
%
%
Specifically, for token selection, we compute the variance statistics of tokens along the channel dimension. If the variance of a token is higher, it tends to contain more useful information for defects. Therefore, we select the top $k$ important tokens by ranking their variances and generating the indexes of these selected tokens.  
Similarly, for channel selection, we first compute the variance statistics of channels along the spatial dimension and then select the top $k$ important channels by sorting the channel variances, generating the indexes of these selected channels.

In the Top-K self-attention module, the Top-K token indexes and channel indexes are first computed on the \textbf{Q}. Then they are used to select the most important tokens and channels from queries \textbf{Q}, generating \textbf{Q}$'$. Meanwhile, they are used to select the most important channels from keys \textbf{K} and the most important tokens from values \textbf{V}, generating \textbf{K}$'$ and \textbf{V}$'$, respectively.
Notably, the Top-K token indexes and channel indexes are only computed on the \textbf{Q} once because the self-attention module first calculates similarities of \textbf{Q} and \textbf{K} and then performs matrix multiplication with \textbf{V}.
If the indexes are re-computed on \textbf{K} and \textbf{V}, the corresponding tokens and channels at the same position in the selected \textbf{Q}, \textbf{K}, and \textbf{V} do not match.
With the obtained \textbf{K}$'$,\textbf{Q}$'$and \textbf{V}$'$, the Top-K self-attention is calculated as follows:
\begin{gather}
    A = Softmax(\frac{\textbf{Q}' \textbf{K}'^T}{\sqrt{C}}) \\
    Z = A\textbf{V}'
\end{gather}
where the $Softmax(.)$ denotes the Softmax function. Compared with the original \textbf{K}, \textbf{Q}, and \textbf{V}, the \textbf{K}$'$, \textbf{Q}$'$and \textbf{V}$'$ have fewer dimensions. So the proposed self-attention can also reduce the computational cost of the self-attention module.

In addition, we introduce a constraint vector to counteract the expansion of unselected regions due to the softmax operation. Because in the Top-K self-attention, each unselected token in the key \textbf{K} also computes similarity with selected tokens in the query \textbf{Q}, generating a score in the attention map. The constraint vectors (6) and (7) are introduced to suppress the scores of these unselected tokens. 
Mathematically, we have:
\begin{gather}
    A_c = Sig(Layernorm(mean(\frac{\textbf{Q}' \textbf{K}'^T}{\sqrt{C}}))) * \gamma \\
    Z' = ZA_c
\end{gather}
where $mean(.)$ is the mean operation, and $Sig(.)$ is the Sigmoid function. $\gamma$ is a learnable parameter.

As stated in ~\cite{yang2022vitkd}, the relations of different tokens and semantic information at shallow layers are weak. Therefore, the proposed Top-K self-attention mechanism is only used in the last two stages of the network. In the ablation experiments, we also observe a performance degradation when using the Top-K self-attention module in the first two stages.

%

\section{Experiments}

\begin{table}[t]
    \setlength\tabcolsep{1.5mm}
    \renewcommand\arraystretch{1.5}
    \centering
     \caption{
    Comparisons of different models on the three defect datasets. It should be noted the CINFormer* denotes that the weights of the ResNet-18 are frozen during the training.
    }
    \label{tab1}
    \begin{tabular}{lcrrcc}
        \hline
        \multicolumn{1}{l|}{} & \multicolumn{1}{c}{}& \multicolumn{1}{c}{}& MT& NEU & DAGM  \\ \cline{4-6} 
        \multicolumn{1}{l|}{\multirow{-2}{*}{Model}} & \multicolumn{1}{c}{\multirow{-2}{*}{\#Param}} & \multicolumn{1}{c}{\multirow{-2}{*}{FLOPs}} & mIoU & mIoU  & mIoU  \\ \hline
        \multicolumn{4}{l}{CNN}  & \multicolumn{1}{l}{} & \multicolumn{1}{l}{} \\ \hline
        \multicolumn{1}{l|}{SPNet [2020]}  & 13M & 1.9G  & 71.8  & 77.6 & 64.5 \\
        \multicolumn{1}{l|}{UNet [2015]}  & 17M   & 3.1G  & 78.7  & 74.7 & 68.2 \\
        \multicolumn{1}{l|}{OCNet  [2021]} & 15M& 11.9G & 80.6 & 75.9  & 68.6   \\
        \multicolumn{1}{l|}{GCAPNet [2020]}  & 67M  & 26.6G & 82.9 & 75.8  &  69.6 \\
        \multicolumn{1}{l|}{ResNet-50 [2016]} & 24M   & 8.2G   & 80.6 & 80.7  & -   \\ \hline
        \multicolumn{4}{l}{Vision Transformer}   & \multicolumn{1}{l}{} & \multicolumn{1}{l}{} \\ \hline
        \multicolumn{1}{l|}{PVT [2021]}   & 20M  & 2.5G & 60.1  & 79.5 & 63.5   \\
        \multicolumn{1}{l|}{SegFormer [2021]}   & 8M  & 21.1G & 68.6 & 82.9& 70.1   \\
        \multicolumn{1}{l|}{DAT [2022]}  & 27M  & 4.6G  & 65.5  & 80.8   & 63.0   \\
        \multicolumn{1}{l|}{Swin-T [2021]} & 28M  & 4.5G   & 62.0 & 80.2   & 70.3   \\ \hline
        \multicolumn{4}{l}{CNN \& Vision Transformer}  & \multicolumn{1}{l}{} & \multicolumn{1}{l}{} \\ \hline
        \multicolumn{1}{l|}{UniFormer [2022]} & 22M & 3.5G  & 58.4 & 80.1   & 70.4  \\
        \multicolumn{1}{l|}{Conformer [2021]}   & 103M   & 23.5G  & 76.5  & 80.9   & 72.4  \\
        \multicolumn{1}{l|}{VST [2021]}  & 44M & 23.2G & -   & 82.8  & 70.1   \\
        \multicolumn{1}{l|}{PGNet [2022]}  & 72M & 17.6G  & 79.1 & 79.9 & 71.2 \\
        \multicolumn{1}{l|}{CINFormer*}   & 30M  & 7.1G  &84.2  &  83.8 & 75.2   \\ 
        \multicolumn{1}{l|}{CINFormer}  & 30M   & 7.1G  & \textbf{86.5} & \textbf{85.7}  & \textbf{78.1}        \\ \hline 
    \end{tabular}

\end{table}

\begin{figure*}[ht]
    \centering
    \includegraphics[width=6.7in]{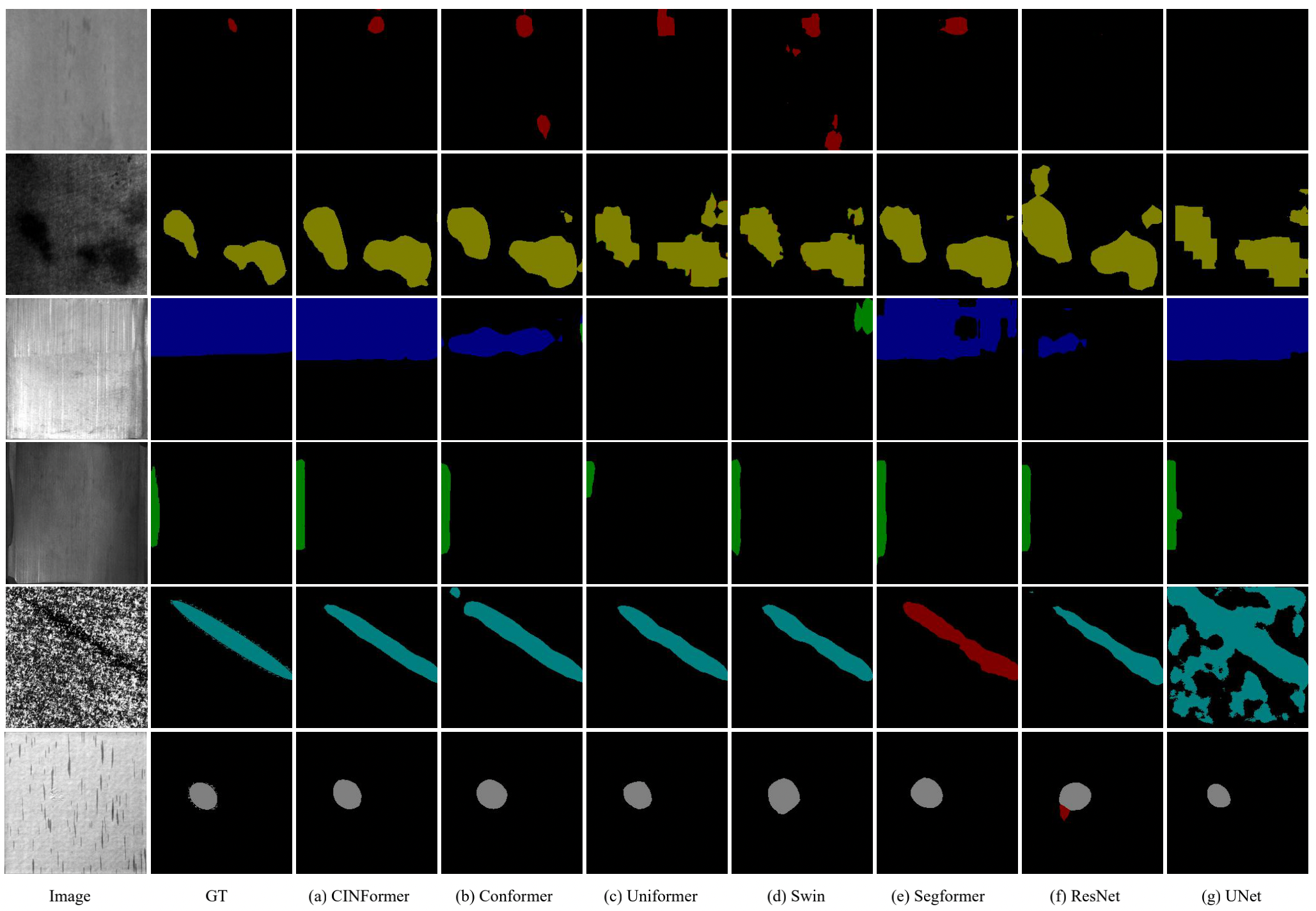}
    \caption{Visualization of segmentation results obtained by various methods. The top two rows are from NEU, the middle two rows are from MT, and the last two rows are from DAGM. Different colors on each dataset represent different kinds of defects. The inconsistent green color in the third row and the inconsistent red color in the fifth and sixth rows present that the methods wrongly recognize the types of these defect areas. And the uncolored image means that the method does not identify a defect.}
    \label{fig4}
\end{figure*} 
\subsection{Datasets}
In the experiment, NEU~\cite{song2013noise}, DAGM 2007 \cite{jager2008weakly}, and  Magnetic tile \cite{huang2020surface} (MT) defect datasets are selected to demonstrate the effectiveness of the proposed method. The details of the three defect datasets are illustrated as follows.

\textbf{NEU} is a steel strip surface defect dataset, which contains three types of defects: patch, inclusion, and scratch.  There are 300 images for each type of defect, and each image is provided with pixel-level ground truth. 
The intra-class difference and inter-class similarity of defects bring great challenges to defect segmentation.

\textbf{DAGM 2007} is artificially generated but the samples are very similar to those in the real world, which contains 10 types of defects generated from various texture and defect models. 
The defect region in each defect image is roughly marked by an ellipse.
Most of the defects are low-contrast and affected by complex background interference, so it is hard to detect them accurately.

\textbf{MT} is a magnetic tile defect dataset, which contains 1344 images. There are 392 defect images, including five categories of defects: uneven, fray, crack, blowhole, and break. These defects exhibit complicated appearances with multi-scale and low contrast, which bring challenges to defect segmentation.

\textbf{Image Size:} The images in the three defect datasets vary from 200$\times$200 to 512$\times$512 in size. It is a common practice to resize the image to a fixed resolution in the defect segmentation task~\cite{dong2019pga, Zhou2022DACNet}.
 \begin{figure*}[t]
    \centering
    \includegraphics[width=6.7in]{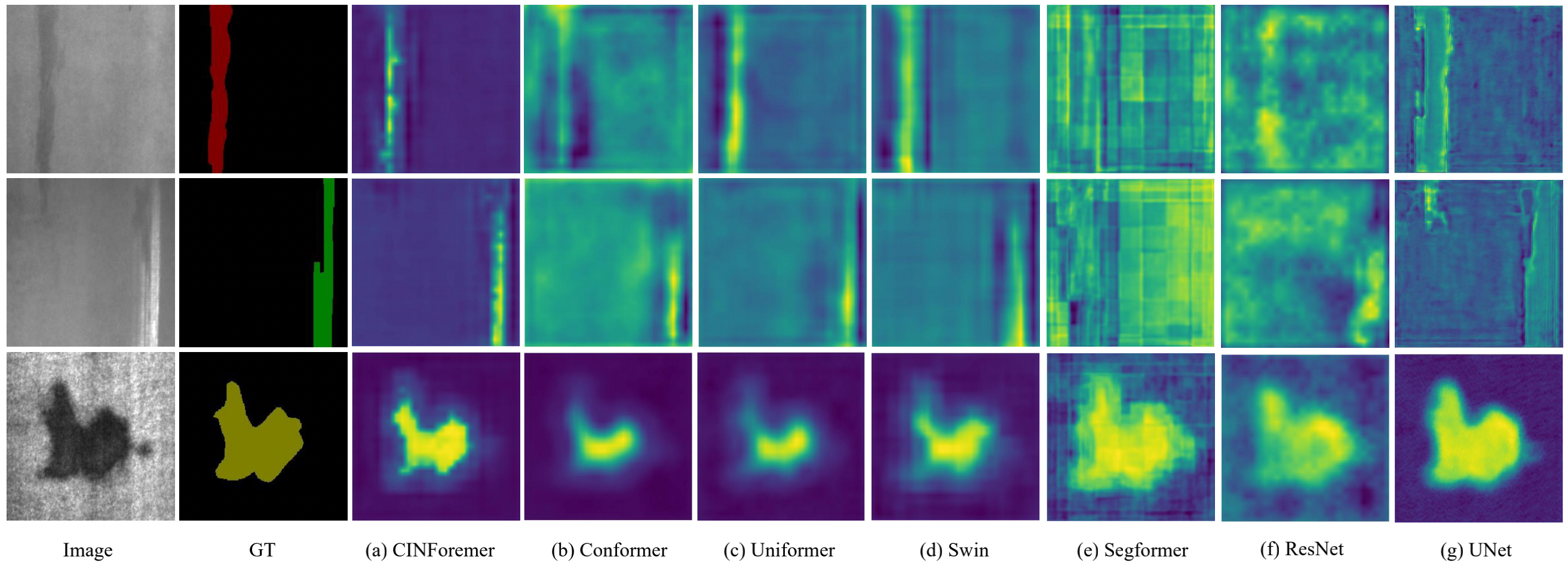}
    \caption{Comparison of feature visualization of different methods. The features are from the last stage of the corresponding models. The yellow color denotes a high response for defect regions while the blue color denotes the background.}
    \label{fig5}
\end{figure*} 

 \subsection{Implementation Details}
We implement the proposed method based on Pytorch~\cite{paszke2017automatic}.
All experiments are conducted on the Tesla V100 platform.
The CNN stem adopts the ResNet-18 pre-trained on the ImageNet~\cite{deng2009imagenet}.  
%
%
We use AdamW~\cite{da2014method} optimizer with a learning rate of 0.00075 and weight decay of 0.005 to train the model.
Meantime, a cosine decay learning rate scheduler is adopted during the training stage. 
For each dataset, 70\% of the images in each category are selected as the training set and the remaining images are used as the test set. Each image is resized to 224$\times$224 during both the training and test stages by following the common practice~\cite{dong2019pga, Zhou2022DACNet}.
With a batch size of 4, the model is trained for 150 epochs, 100 epochs, and 100 epochs on NEU, DAGM 2007, and MT defect datasets, respectively, where the first 20 epochs are used for warming up. In the experiment, the mean Intersection of Union (mIoU) is adopted to evaluate the performance of different models. We utilize the cross-entropy loss as the supervision to train the network.

\begin{figure}[t]
    \centering
    \includegraphics[width=\linewidth]{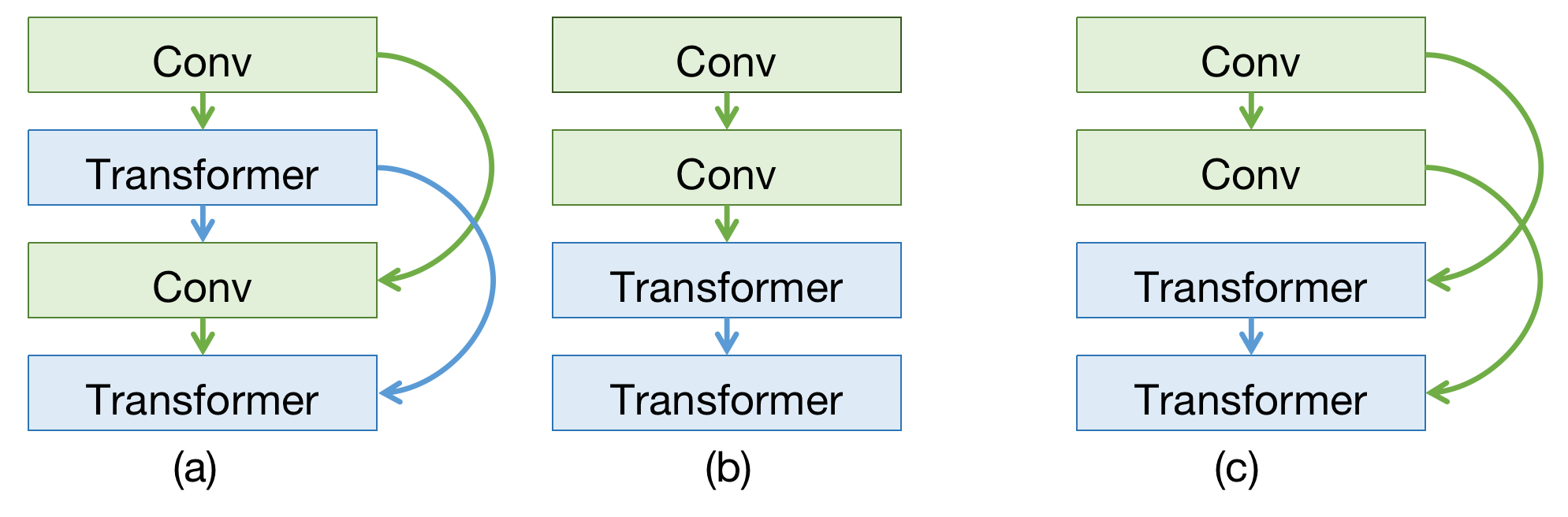}
    \caption{The illustration of different CNN injection structures. (a)$\sim$(c) represent the structures of bidirectional feature interaction, the structure of convolution followed by transformer, and the proposed multi-stage CNN feature injection structure, respectively.}
    \label{fig6}
\end{figure}

\subsection{Comparison with State-of-the-arts}
The proposed CINFormer is compared with 13 state-of-the-art models on DAGM 2007, NEU, and MT defect datasets. 
These methods are classified as CNN-based methods including UNet~\cite{ronneberger2015u}, ResNet~\cite{he2016deep}, SPNet~\cite{hou2020strip}, GCAPNet~\cite{chen2020global}, and OCNet~\cite{yuan2021ocnet}, vision transformer based methods including PVT~\cite{wang2021pyramid}, Segformer~\cite{xie2021segformer}, Swin-T~\cite{liu2021swin}, and DAT~\cite{xia2022vision}, and methods of combining CNN and transformer including Uniformer~\cite{li2022uniformer}, Conformer~\cite{peng2021conformer}, VST~\cite{liu2021visual}, PGNet~\cite{xie2022pyramid}.

The comparison results of various methods are given in Table~\ref{tab1}. It is seen that CNN-based methods are superior to transformer-based methods on MT. This is because there are many weak defects on the MT dataset and CNN is better at capturing defect details than the transformer. 
On the contrary, transformer-based methods outperform CNN-based methods on NEU. This is because the most of defects have relatively complex background disturbances. This indicates that the transformer models can better suppress the background interference in complex scenes than CNN. 
On the DAGM dataset, the transformer-based methods and the CNN-based methods perform similarly. 
As to the existing hybrid methods of combining CNN and transformer, they perform differently on three datasets. They generally perform worse than CNN on MT, perform almost the same as the transformer on NEU, and perform a little better on DAGM. This means the existing hybrid methods do not effectively exploit their respective merits on the surface defect datasets.
The proposed CINFormer outperforms the above methods on three defect datasets. 
It is noted that the CINFormer can only fine-tune the Transformer part while keeping the weights of the ResNet-18 frozen during training. Even so, CINFormer outperforms Conformer on the defect datasets. When trained in an end-to-end way, CINFormer obtains the best result on three defect datasets.
In general, the experimental results demonstrate the effectiveness of the proposed one-way CNN injection. 

Fig. \ref{fig4} presents some segmentation results of various methods. It is found that some methods fail to detect small or low-contrast defects, such as ResNet, UNet. Some methods such as Segformer and Swin even misclassify the defects, manifested by inconsistent colors between predictions and ground truth (GT) on defect regions in Fig. \ref{fig4}. On the contrary, the proposed CINFormer can accurately detect those defect regions.
In addition, Fig. \ref{fig5} presents the feature visualization results of various models on three defect samples. It is found that our method can better focus on defect regions while suppressing background interference.

\subsection{Ablation Study}
To further demonstrate the effectiveness of the proposed CNN injection strategy and the Top-K self-attention module, we conduct the following ablation experiments.

\textbf{Ablation on the Transformer with CNN Injection:}
To demonstrate the priority of the proposed CNN feature injection manner, we conduct several experiments based on the models with and without the Top-K self-attention module. We first compare the Swin-T model with the CINFormer model without the Top-K self-attention module denoted as CINFormer (w/o) and then compare the two models having the Top-K self-attention module. The Swin-T model with the Top-K self-attention module is denoted as Swin-T (Top-K).
The experimental results in Table \ref{tab2} show that CINFormer outperforms the Swin-T in both conditions. CINFormer (w/o) surpasses Swin-T by 6.5 points and CINFormer surpasses Swin-T (Top-K) by 7.0 points in terms of mIoU on DAGM. 
This indicates that the injected multi-level CNN features are beneficial for the transformer to detect defects.

Furthermore, the proposed multi-stage CNN injection manner is compared with other 
hybrid structures of CNN and transformer, as shown in Fig. \ref{fig6} (a) and (b). 
As shown in Table \ref{tab3}, the proposed method obtains improvements of 1.0 and 1.8 points over the other two structures, respectively. 
It is noted that structure (a) is a bidirectional feature injection pattern with features of CNN and transformer interacting with each other, which is similar to Conformer~\cite{peng2021conformer}. 
Though this kind of bidirectional interaction can promote the full merging of CNN features and transformer features, it impairs the ability of the CNN branch to represent the detailed features. 
As a result, our proposed one-way CNN feature injection method outperforms it on the defect dataset.

\begin{table}[t]
    \setlength\tabcolsep{1.5mm}
    \renewcommand\arraystretch{1.5}
    \centering
     \caption{
    Comparison of the CINFormer (w/o) and Swin-T on the DAGM dataset. CINFormer (w/o) denotes the CINFormer without the Top-K self-attention module. Swin (Top-K) denotes the Swin transformer with the
    Top-K Self-attention.
    }
    \label{tab2}
    \begin{tabular}{l|crcc}
        \hline
         & \multicolumn{1}{l}{}   & \multicolumn{1}{l}{}  & mIoU \\ 
         \cline{4-4} 
        \multirow{-2}{*}{Model}   & \multicolumn{1}{l}{\multirow{-2}{*}{\#Prama}} & \multicolumn{1}{l}{\multirow{-2}{*}{FLOPs}} & DAGM \\ 
        \hline
        \multicolumn{1}{l|}{Swin-T} & 28M   & 4.5G  & 70.3 \\
        \multicolumn{1}{l|}{CINFormer (w/o)} & 32M& 7.3G   & 76.8 \\ 
        \multicolumn{1}{l|}{Swin-T (Top-K)}  & 26M  & 4.5G & 71.1 \\
        \multicolumn{1}{l|}{CINFormer} & 30M & 7.1G & \textbf{78.1} \\
        \hline
    \end{tabular}
   
\end{table}

\begin{table}[t] 
    \caption{
    Comparison of different CNN feature injection structures presented in Fig. \ref{fig6} on the DAGM dataset. 
    }
    \label{tab3}
     \setlength\tabcolsep{1.5mm}
    \renewcommand\arraystretch{1.5}
    \centering
    \begin{tabular}{r|crccc}
        \hline
        & \multicolumn{1}{l}{} & \multicolumn{1}{l}{} & \multicolumn{1}{l}{}& mIoU \\ 
        \cline{5-5} 
        \multirow{-2}{*}{Structure} & \multicolumn{1}{l}{\multirow{-2}{*}{\#Prama}}  & \multicolumn{1}{l}{\multirow{-2}{*}{FLOPs}} & \multicolumn{1}{l}{\multirow{-2}{*}{Time/epoch}} & DAGM \\ 
        \hline
        \multicolumn{1}{c|}{(a)} & 28M  & 6.7G   & 102s  & 75.8 \\
        \multicolumn{1}{c|}{\cellcolor[HTML]{FFFFFF}{\color[HTML]{000000}(b)}}  & \cellcolor[HTML]{FFFFFF}{\color[HTML]{000000} 29M} & \cellcolor[HTML]{FFFFFF}7.1G    & 101s   & 75.0 \\
        \multicolumn{1}{c|}{(c)}   & 30M & 7.3G & 95s & \textbf{76.8} \\ 
        \hline
    \end{tabular}
\end{table}

%
 \begin{table}[t]
    \centering
    \caption{
    Comparison of the proposed Top-K self-attention module and the traditional self-attention module in different transformer models. CINFormer(w/o) denotes the CINFormer without the Top-K self-attention module. 
    }
    \label{tab4}
    \setlength\tabcolsep{1.5mm}
    \renewcommand\arraystretch{1.5}
    \begin{tabular}{l|crrccc}
        \hline
        & \multicolumn{1}{c}{} & \multicolumn{1}{c}{} & MT & NEU  & DAGM  \\ \cline{4-6} 
        \multirow{-2}{*}{Model}  & \multirow{-2}{*}{\#Param} & \multicolumn{1}{c}{\multirow{-2}{*}{FLOPs}} & mIoU  & mIoU   & mIoU  \\ 
        \hline
        Swin-T [2021]   & 28M  & 4.5G & 64.0 & 80.0 & 70.3   \\ 
        Swin-T (Top-K) & 26M   & 4.3G  & \textbf{66.7} & \textbf{81.6} & \textbf{71.1} \\ 
        \hline
        Conformer [2021]   & 103M  & 23.6G  & 76.5 & 80.9  & 71.6  \\ 
        Conformer (Top-K)  & 103M  & 23.0G  & \textbf{76.9} & \textbf{81.2} & \textbf{72.4} \\ 
        \hline
        Uniformer [2022]  & 21M   & 3.6G  & 58.2  & 80.1  & 69.7   \\
        Uniformer (Top-K)  & 20M & 3.4G  & \textbf{60.2} & \textbf{80.6} & \textbf{70.4} \\ 
        \hline              
        CINFormer (w/o)   & 32M   & 7.3G  & 85.6 & 85.2  & 76.8  \\
        CINFormer & 30M  & 7.1G   & \textbf{86.5} & \textbf{85.7} & \textbf{78.1} \\ 
        \hline
\end{tabular}
\end{table}

\begin{table}[t]   
\setlength\tabcolsep{1.5mm}
\renewcommand\arraystretch{1.5}
\centering
\caption{
Comparison of using Top-K self-attention at different stages on the DAGM dataset. It should be noted that “$\checkmark$” denotes self-attention module is replaced with Top-K self-attention in stage i.
}
\label{tab5}
\begin{tabular}{c|cccc|ccc}
\hline

\multirow{2}{*}{Model} & \multicolumn{4}{c|}{Stage} & \multirow{2}{*}{\#Params} & \multirow{2}{*}{FLOPs} & DAGM \\
\cline{2-5} \cline{8-8}
~& 1  & 2  & 3  & 4   &   ~ &   ~ & mIoU \\
\hline
\multirow{5}{*}{Swin-T (Top-K)}   &&&&  &27M &4.4G &70.3 \\
~ &&&&\checkmark  &27M  &4.3G  &70.5    \\
~ &&&\checkmark &\checkmark     &26M    &4.3G  &\textbf{71.1}                          \\
~ &&\checkmark &\checkmark &\checkmark  &26M    &4.3G    &70.7                           \\
\multicolumn{1}{l|}{} &\checkmark &\checkmark &\checkmark & \checkmark   &25M         &4.2G       &70.8              \\ 
\hline

\multirow{5}{*}{CINFormer}  &&&&   &31M  &7.2G &77.9  \\
~ &&&&\checkmark   &31M &7.2G&\textbf{78.1}    \\
~ &&&\checkmark &\checkmark  &30M &7.1G &\textbf{78.1}    \\
~ &&\checkmark &\checkmark &\checkmark  &30M  &7.1G   &77.9   \\
~&\checkmark &\checkmark &\checkmark & \checkmark  &29M &7.0G  &77.8  \\ 
\hline

\end{tabular}
\end{table}
 
\begin{figure}[t]
    \centering
    \includegraphics[width=3.2in]{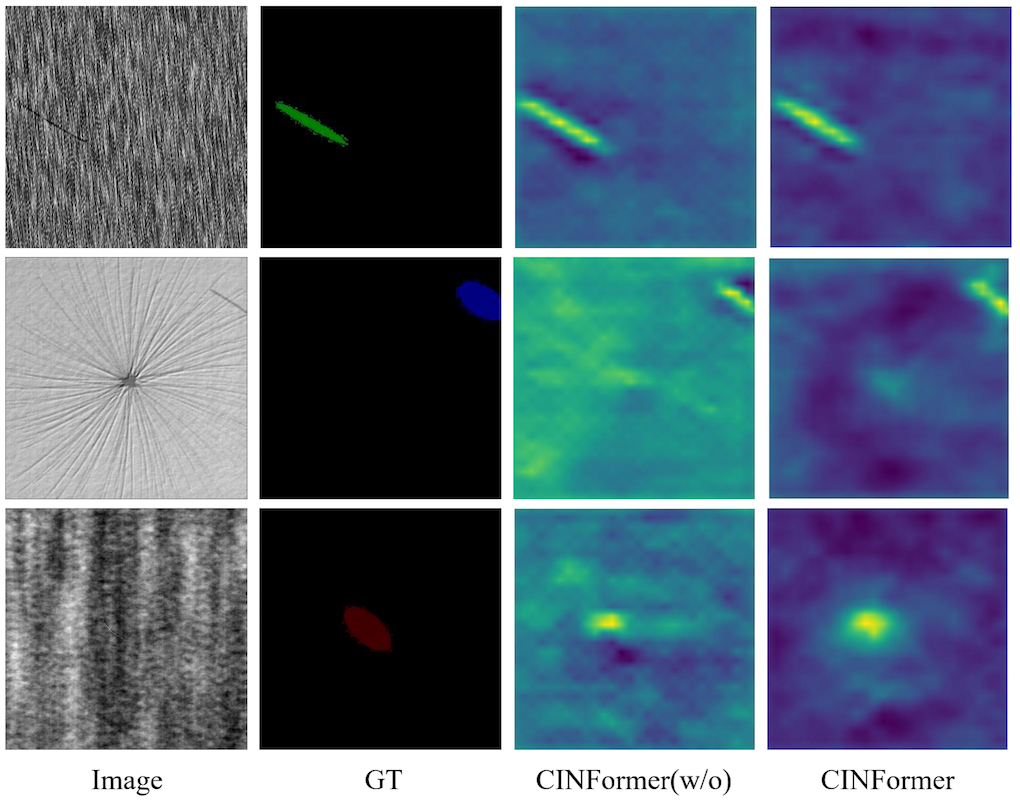}
    \caption{Comparison of the feature visualization of CINFormer and CINFormer without Top-K self-attention module denoted as CINFormer (w/o).}
    \label{fig7}
\end{figure}

\textbf{Ablation on the Top-K Self-attention Module:}
To demonstrate the effectiveness of the proposed Top-K self-attention module in the transformer network, we use it to replace the self-attention module of different transformer models. 
Specifically, we only replace the self-attention module in the last two stages of the transformer models. The experimental results in Table~\ref{tab4} show that the introduction of Top-K self-attention module improves mIoU by 0.8, 0.8, 0.7, and 1.3 points compared with Swin-T, Conformer, Uniformer, and CINFormer (w/o) on the DAGM defect dataset, respectively. 
This indicates that the proposed Top-K self-attention module is a plug-and-play module, which can improve the performance of various transformer models in defect detection.
Meanwhile, the proposed Top-K self-attention module reduces the computational complexity of the transformer model because it only selects important tokens and channels to calculate attention. 
Furthermore, we present the feature visualization of CINFormer and CINFormer (w/o), as shown in Fig. \ref{fig7}. The introduction of the Top-K self-attention module can make CINFormer focus on defect regions and suppress background interference.  

\textbf{Configuration of the Top-K Self-attention Module:}
First, we compare the performance of models using Top-K self-attention at different stages. To be specific, experiments are conducted based on the models with and without the CNN injection module, as shown in Table~\ref{tab5}. The Swin-T model contains four stages, denoted as 1, 2, 3, and 4, respectively. We gradually add the proposed Top-K self-attention in the transformer network starting from the last stage.
The proposed CINFormer achieves the best performance when the Top-K self-attention is used in the last two stages. This is because a single token in the earlier stage is not highly correlated with its surrounding tokens~\cite{yang2022vitkd}, which makes it hard to reconstruct the abandoned tokens. Therefore, using the Top-K self-attention mechanism in the early stage will make the model lose some important detailed information during feature extraction, thereby reducing the accuracy of recognition.
In contrast, a single token in the latter stage has a strong correlation with its surrounding tokens, the defect (target) tokens are easily affected by background noise. 
%
To sum up, when Top-K self-attention modules are used in the last two stages,  the proposed CINFormer can achieve the best balance between accuracy and cost.


%
Secondly, we further investigate the effect of the hyperparameter K on model performance. The experiments are conducted on the model under the CNN-injected module and the non-CNN-injected module.
The K is set to 14, 21, 28, 35, and 42, respectively.
As shown in Table~\ref{tab6}, the proposed CINFormer achieves the best performance when K is set to 28. 
Because the K with small values will discard a lot of tokens, including some useful information.
And the K with large values will overwhelm the detailed information, which is the same as the original self-attention mechanism.

\begin{table}[t] 
  
\setlength\tabcolsep{1.5mm}
\renewcommand\arraystretch{1.5}
\centering
 \caption{
Comparison of different K values of the Top-k self-attention module on the DAGM dataset. It should be noted Swin (Top-K) denotes the Swin transformer with the Top-K Self-attention. 
}

\label{tab6}
\begin{tabular}{c|cccc}
\hline
&\multicolumn{1}{c}{}  &\multicolumn{1}{c}{}  &\multicolumn{1}{c}{}   & DAGM   \\ 
\cline{5-5} 
\multirow{-2}{*}{Model} & \multirow{-2}{*}{Number-K} & \multirow{-2}{*}{\#Param} & \multirow{-2}{*}{FLOPs} & \multicolumn{1}{c}{mIoU} \\ 
\hline
\multicolumn{1}{c|}{}  &14  &25M &4.2G  &70.7  \\
\multicolumn{1}{l|}{}  &21 &26M  &4.3G &70.8    \\
\multicolumn{1}{l|}{Swin-T (Top-K)} &28  &26M  &4.3G &\textbf{71.1} \\
\multicolumn{1}{l|}{} &35 &27M  &4.5G &71.0   \\
\multicolumn{1}{l|}{} &42 &27M  &4.5G &71.0  \\ 
\hline
\multicolumn{1}{l|}{} &14 &30M  &7.0G &77.3  \\
\multicolumn{1}{l|}{} &21 &30M &7.0G &77.6 \\
\multicolumn{1}{c|}{CINFormer} &28 &30M&7.1G  &\textbf{78.1} \\
\multicolumn{1}{l|}{}  &35 &31M &7.1G  &78.0   \\
\multicolumn{1}{l|}{}  &42 &31M  &7.2G &77.0  \\ 
\hline
\end{tabular}
   
\end{table}

\section{Conclusion}
In this paper, we propose a novel transformer network with multi-stage CNN feature injection for surface defect segmentation, termed CINFormer. CINFormer is a UNet-like architecture in which the encoder takes advantage of the transformer and the multi-level CNN features to promote the representation capacity of the feature. In the meantime, a Top-K self-attention module is presented to reduce the impact of redundant background information. It selects valuable tokens and channels to guide the model to highlight defect regions. The extensive experiments on DAGM 2007, Magnetic tile, and NEU defect datasets demonstrate the effectiveness of the proposed CINFormer in different defect scenes.
 

\bibliographystyle{ieeetr}
\bibliography{Tycb}

\end{document}